# LR-Net: A Lightweight and Robust Network for Infrared Small Target Detection

Chuang Yu[1,2,3,4], Yunpeng Liu[1, 2, *], Jinmiao Zhao[1,2,3,4], and Zelin Shi[1, 2]

[1]Key Laboratory of Opto-Electronic Information Processing, Chinese Academy of Sciences
[2]Shenyang Institute of Automation, Chinese Academy of Sciences
[3]Institutes for Robotics and Intelligent Manufacturing, Chinese Academy of Sciences
[4]University of Chinese Academy of Sciences
{yuchuang, ypliu, zhaojinmiao, zlshi}@sia.cn

**Abstract.** Limited by equipment limitations and the lack of target intrinsic features, existing infrared small target detection methods have difficulty meeting actual comprehensive performance requirements. Therefore, we propose an innovative lightweight and robust network (LR-Net), which abandons the complex structure and achieves an effective balance between detection accuracy and resource consumption. Specifically, to ensure the lightweight and robustness, on the one hand, we construct a lightweight feature extraction attention (LFEA) module, which can fully extract target features and strengthen information interaction across channels. On the other hand, we construct a simple refined feature transfer (RFT) module. Compared with direct cross-layer connections, the RFT module can improve the network's feature refinement extraction capability with little resource consumption. Meanwhile, to solve the problem of small target loss in high-level feature maps, on the one hand, we propose a low-level feature distribution (LFD) strategy to use low-level features to supplement the information of high-level features. On the other hand, we introduce an efficient simplified bilinear interpolation attention module (SBAM) to promote the guidance constraints of low-level features on high-level features and the fusion of the two. In addition, We abandon the traditional resizing method and adopt a new training and inference cropping strategy, which is more robust to datasets with multi-scale samples. Extensive experimental results show that our LR-Net achieves state-of-the-art (SOTA) performance. Notably, on the basis of the proposed LR-Net, we achieve 3rd place in the "ICPR 2024 Resource-Limited Infrared Small Target Detection Challenge Track 2: Lightweight Infrared Small Target Detection".

## 1  Introduction

With the rapid development of infrared imaging and digital image processing technology, infrared small target detection technology has been widely used in fields such as maritime surveillance, early warning and precision guidance [1, 2]. Owing to the lack of target intrinsic features and the limitation of network size in practical applications, building a high-precision, lightweight and robust infrared small target detection network is highly important.

Early research on infrared small target detection focused mainly on non-deep learning-based methods, including background suppression-based methods [3-5], human visual system-based methods [6-8], and image data structure-based methods [9-11]. Although these methods have excellent detection effects in specific scenarios, when the scenarios are diverse and complex, the above methods have many false detections or even fail. Considering that deep learning networks have powerful feature extraction capabilities, subsequent research on infrared small target detection has focused mainly on deep learning methods. For example, GGL-Net [12] focuses on using the edge information of infrared small targets to guide the network to extract the target in a refined manner. MSDA-Net [13] focuses on extracting high-frequency directional features to achieve effective and sufficient extraction of small target features. However, existing infrared small target detection methods often focus on detection accuracy and ignore the lightweight of the network.

To solve this problem, we aim to build a high-precision, lightweight and robust infrared small target detection network. First, considering that an excellent feature extraction part is crucial to the entire network, we explore it in depth and construct a lightweight feature extraction attention (LFEA) module. The LFEA module achieves full and effective feature extraction through a dual-branch path, and uses the ECA module [14] to further extract the relationships between channels. Secondly, since infrared small targets are small and weak, there is a risk of losing target features in high-level feature maps. To solve this problem, on the one hand, we propose a low-level feature distribution (LFD) strategy, which distributes the low-level feature maps step by step to the high-level feature maps in the feature extraction stage to supplement the information. On the other hand, we introduce an efficient simplified bilinear interpolation attention module (SBAM) [15] to promote the guidance constraints of low-level features on high-level features and the fusion of the two, thereby reducing the loss of target information. Thirdly, considering that the direct cross-layer connections in U-Net [16] may result in imprecise features being transferred, we construct a simple refined feature transfer (RFT) module to improve the network's feature extraction and transfer capabilities with little resource consumption. Finally, to solve the problem of multi-size samples, we abandon the traditional method of directly resizing the image to a fixed size and adopt a new cropping method. During training phase, we randomly crop samples to 256 × 256 pixels. For testing phase, we use sliding window cropping method. Compared with the direct resizing method, the cropping method can learn more samples and avoid target loss after resizing due to the sample size being too large.

In summary, we propose an innovative lightweight and robust network (LR-Net), which abandons the complex structure and achieves an effective balance between detection accuracy and resource consumption. Our code can be available online[1]. On the basis of the proposed LR-Net, we won third place in "ICPR 2024 Resource-Limited Infrared Small Target Detection Challenge Track 2: Lightweight Infrared Small Target Detection". The contributions of this manuscript can be summarized as follows:

---

[1] https://github.com/YuChuang1205/LR-Net

(1) To ensure the lightweight and robustness of the constructed network, on the one hand, we construct an LFEA module to effectively extract target features and strengthen information interaction across channels. On the other hand, we construct a simple RFT module to further improve the network's feature refinement extraction and transfer capabilities.

(2) Infrared small targets are small and weak, and their features are easily submerged as the network deepens. To solve this problem, on the one hand, we propose an LFD strategy, which distributes low-level features to each high-level feature map. On the other hand, we introduce efficient SBAM to promote the fine fusion of high-level and low-level feature maps to avoid information loss.

(3) To solve the problem of multi-scale samples, we abandon the traditional method of directly resizing images to a fixed size and adopt a new training and inference cropping strategy, which has higher detection accuracy and robustness.

## 2  Related Work

### 2.1  Non-deep learning-based infrared small target detection methods

Early infrared small target detection methods are mainly non-deep learning-based methods, which can be divided into three main categories: background suppression-based methods [3-5], human visual system-based methods [6-8] and image data structure-based methods [9-11]. Background suppression-based methods mainly highlight the target by reducing background clutter, such as Max-Median filter [3], bilateral filter [4] and Top-Hat filter [5]. Human visual system-based methods imitate the characteristics of the human visual system and improve the target detection rate by enhancing the salience area of the image, such as the local contrast metric (LCM) [6], multi-scale local contrast metric (MLCM) [7] and weighted local difference measure (WLDM) [8]. Image data structure-based methods transform the small target detection problem into an optimization problem of low-rank and sparse matrix recovery, for example, the infrared image patch model (IPI) [9], the minimum joint L2,1 normal form (NRAM) [10], and non-negative infrared patch-image model based on partial sum minimization of singular values (NIPPS) [11]. Although the above methods have excellent performance in specific scenarios, they require careful model design and hyperparameter tuning. Moreover, their detection capabilities need to be improved, especially for diverse and complex scenarios.

### 2.2  Deep learning-based infrared small target detection methods

Compared with model-driven non-deep learning methods, convolutional neural networks (CNNs) can automatically learn the nonlinear mapping between input images and true labels in a data-driven manner. This enables the deep learning-based infrared small target detection method to adapt more effectively and extend to complex practical scenarios. In the early days, Dai et al. propose an asymmetric context module [17] and further add a local contrast metric [18] to improve the target contrast. To further enhance the extraction of the relationships among local features,

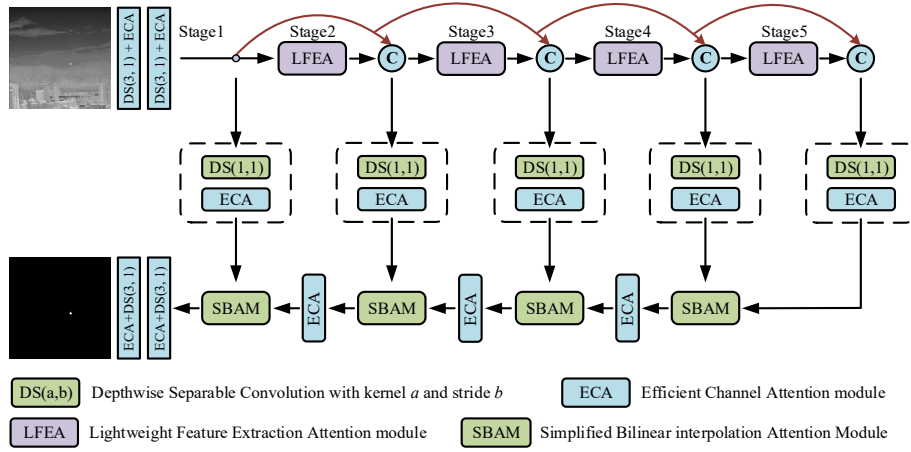

**Fig. 1.** Network structure of LR-Net. The red line denotes the max pooling operation with a kernel of 2 and a stride of 2. The dotted area denotes the RFT module.

Yu et al. successively construct a multi-scale local contrast learning network (MLCL-Net) [19] and an attention-based local contrast learning network (ALCL-Net) [15]. Subsequently, Zhao et al. further attempt to introduce prior knowledge in the field of infrared small targets (edge and high-frequency directional features) into the network, and successively propose GGL-Net[12] and MSDA-Net[13]. In addition, Zhang et al. propose an ISNet [20] that focuses on shape features, which enhances the detection performance by extracting and utilizing the shape information of the target. Li et al. propose an DNANet [21] that combines dense connections and an attention mechanism to capture and highlight the features of small targets effectively while reducing the influence of background clutter. Wu et al. propose a U-Net nested network (UIU-Net) [22] to achieve multi-scale feature fusion and more refined segmentation results. Although the above methods have excellent infrared small target detection accuracy, they ignore the lightweight requirement. The resource consumption of the above methods is significant, and it is often difficult to meet actual comprehensive performance requirements. Therefore, we aim to build a high-precision, lightweight and robust infrared small target detection network.

## 3      Methods

### 3.1     LR-Net

Considering that infrared small target detection is often used in military, surveillance, drone and other fields, these applications usually require real-time processing capabilities. Therefore, we construct a lightweight and robust network (LR-Net) for infrared small target detection. We abandon the complex structure and achieve an effective balance between accuracy and resource consumption. Fig. 1 shows that LR-Net can be divided into three parts: feature extraction (top), feature transfer (middle) and feature fusion (bottom). In the feature extraction part, on the one hand, we use our proposed LFEA module, which can fully extract effective features and strengthen the interaction between feature channels while ensuring lightweight. By superimposing the constructed LFEA modules, a 5-stage basic feature extraction network structure is formed. On the other hand, considering the weak and small characteristics of infrared small targets, there is a risk that the targets will be submerged by the background after the original image is downsampled multiple times. Therefore, we use our proposed LFD strategy, which downsamples the underlying features through the max pooling operation and then concatenates them with the high-level feature map in the feature channel dimension. Notably, the channel dimensions of Stages 1-5 are 4, 8, 16, 32, and 64. In the feature transfer part, to improve the effectiveness of refined information transfer and ensure lightweight, different from direct feature transfer, we use a layer of depthwise separable (DS) convolution and an ECA layer. For convenience, we call this combined structure the RFT module. Considering that the RFT module is very simple, we do not detail it in the following sections. In the feature fusion part, we introduce the SBAM proposed in our previous work [15]. This module uses low-level features to perform fine-grained attention constraints on high-level features and fuses the low-level features with the constrained high-level features. Compared with the multi-level feature fusion strategy of the FPN [23], this module not only has fewer parameters but also promotes the fine fusion of high-level features and low-level features. To further improve the effectiveness of the output features at each stage, we add an ECA module after each SBAM. In addition, it is worth mentioning that during the network training and inference process, we abandon the traditional method of directly resizing the image to a fixed size and adopt the cropping method. During training, we randomly crop the samples to 256×256 pixels. During the inference process, we use a sliding window cropping method instead of directly inputting the original image or resizing it.

### 3.2 Lightweight feature extraction attention module

To ensure the robustness and lightweight of the constructed network, we propose a lightweight feature extraction attention (LFEA) module, which is simple and efficient. From Fig. 2, first, to ensure the lightweight of the network, we use depthwise separable convolution instead of ordinary convolution. Compared with ordinary convolution operations, depthwise separable convolution decomposes it into two independent operations: depthwise convolution and point-by-point convolution. It greatly reduces the number of parameters and computational complexity while ensuring network accuracy. Secondly, unlike using depthwise separable convolution alone, we construct a new branch to supplement information. Considering the small size of infrared small targets, using depthwise separable convolution with a stride of 2 will cause some information loss. Therefore, on the basis of prior knowledge that infrared small targets are generally bright areas in an image, we cleverly use a depthwise separable convolution with a stride of 1 and combine it with a max pooling operation to supplement additional information. It greatly improves the feature extraction capability of the network while introducing a small number of parameters.

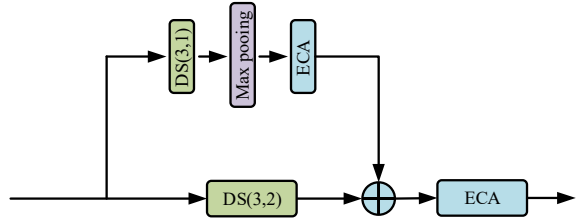

**Fig. 2.** Structure of the LFEA module. The meaning of each part is the same as before.

Finally, to further strengthen the modeling of the relationships between the channels and improve the refinement of the generated feature maps, we introduce the ECA module and reasonably embed it into the LFEA module. Notably, ECA is a very lightweight attention mechanism. In summary, the LFEA module is a simple and efficient feature extraction module. By superimposing this module, an efficient feature extraction network for infrared small target detection can be constructed.

### 3.3 Low-level feature distribution strategy

Infrared small targets are characterized by small sizes and weak intensities. Simply performing convolutional superposition increases the risk of target features being submerged by background features in the deep feature map. To solve this problem and take lightweight into consideration, we propose a low-level feature distribution (LFD) strategy, the detailed structure of which is shown in the red line in Fig. 1. To solve the problem of infrared small target feature loss in high-level feature maps, we think it would be effective to use low-level features to supplement information. Considering the fact that infrared small targets are generally distributed in brighter areas of the image, the ability of low-level features to better preserve the target features, and the need for lightweighting, we directly perform max pooling operations on the feature map output from the first stage to supplement the information of each high-level feature map, so as to solve the problem of high-level target loss at a low cost. Specifically, we perform multiple max pooling operations on the feature maps of the first stage to meet the splicing size requirement of each stage. Max pooling can better preserve the feature information of infrared small targets without introducing additional parameters during the downsampling process of the feature map.

### 3.4 Training and inference cropping strategies

Considering the multi-size challenges in actual scenes, for large images, directly adopting the strategy of resizing to a fixed size will cause the size of the target in the image to decrease and become difficult to detect. To address this problem, we abandon the traditional resizing method and use a training and inference cropping strategy. From Fig. 3, during the training phase, we randomly crop the specified size from the original image for network training. This process allows the network to learn more samples, which can alleviate the problem of infrared small target samples to a certain extent. In the inference phase, we use a sliding window cropping method to slide the entire image into a window of a specified size. For the part that is not larger than the specified window, black edges are added, and then the original image segmentation area is cropped after stitching. This method can avoid the problem of a reduced detection rate due to resizing because the original image is too large. In addition, when the network training image is 256×256 pixels, the inference accuracy is reduced if the test image size is inconsistent with the training.

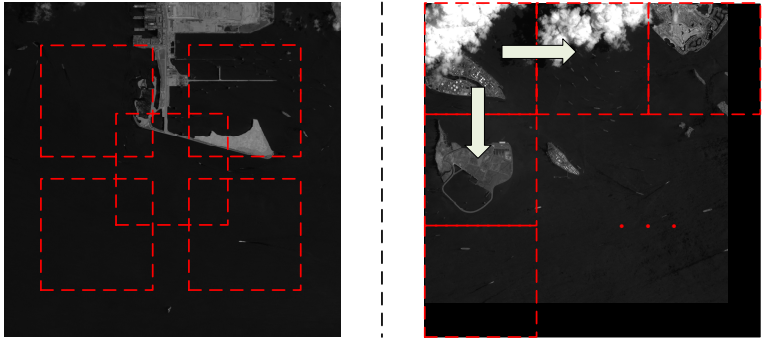

**Fig. 3.** Training and inference cropping strategies. The left side shows random cropping during the training phase, and the right side shows window sliding cropping during the inference phase.

### 3.5 Loss Function and Network Optimization

To refine the extraction of the infrared small target area, we pay extra attention to the edges of the infrared small target in the construction of the loss function. We use edge enhancement loss as our final loss function for network optimization, which encourages the network to pay more attention to the edge of the target during the optimization process, thereby achieving fine detection of the target area. It is expressed as follows:

$$L_{EE}(p,t,w) = \frac{1}{N}\sum_{i=1}^{N} w_i \cdot BCE(p_i,t_i) \quad (1)$$

$$w_i = \begin{cases} 4 & \text{if pixel } i \text{ is at the edge} \\ 1 & \text{otherwise} \end{cases} \quad (2)$$

$$BCE(p,t) = -\frac{1}{N}\sum_{i=1}^{N}[t_i \log(\sigma(p_i)) + (1-t_i)\log(1-\sigma(p_i))] \quad (3)$$

where $P_i$ denotes the predicted probability that the point belongs to the target. $t_i$ denotes the true label. $BCE$ denotes the binary cross-entropy loss.

## 4 Experiments

### 4.1 Dataset

Consistent with "ICPR 2024 Resource-Limited Infrared Small Target Detection Challenge Track 2: Lightweight Infrared Small Target Detection", we use seven public datasets, SIRST-V2, IRSTD-1K, IRDST, NUDT-SIRST, NUDT-SIRST- sea, NUDT-MIRSDT, and Anti-UAV410, totaling 9,000 images, as experimental datasets. The images collected in this dataset have multiple observation perspectives (land-based, air-based, and space-based), multiple target types (extended targets, spot targets, and point targets), multiple bands (shortwave, longwave, and near-infrared), and multiple resolutions (256×256, 512×512, 1024×1024, etc.). For the convenience of description and considering its characteristics, we call the dataset the wide-area infrared small target detection (W-IRSTD) dataset. The W-IRSTD dataset is extremely challenging because it combines multiple public datasets. Fig. 4 shows some samples from the W-IRSTD dataset. We can see that the differences between the samples are very obvious and that the types of objects are diverse. In the experiment, we divide the 9,000 images into training and test sets at a 4:1 ratio.

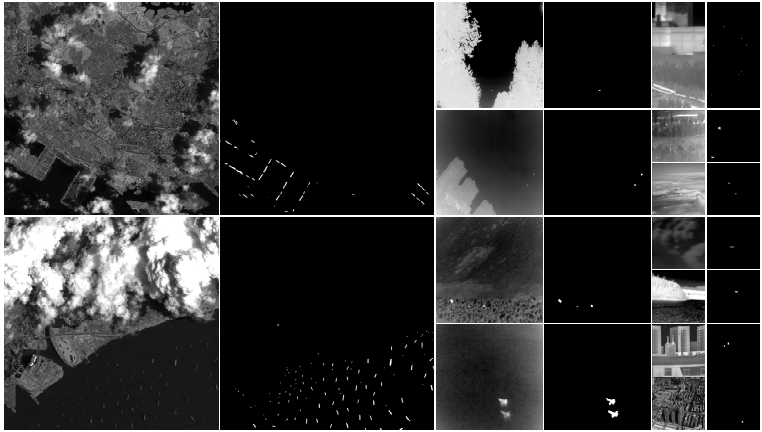

**Fig. 4**. Some samples from the W-IRSTD dataset and their true labels.

### 4.2  Experimental settings

**Experimental environment and parameter settings.** The experimental environment is an Ubuntu 18.04 operating system, and the GPU is an RTX 3090 24 GB. The training epoch, batch size, and learning rate are 300, 10, and 5e-4, respectively. The batch size for the test is 1. The default window size for both training and testing is 256×256 pixels.

**Data augmentation.** Deep learning methods usually require many training samples to learn. Therefore, we perform random data augmentation operations on the training samples before inputting them into the network. For example, flipping, rotation, contrast adjustment, etc.

**Evaluation metrics.** Consistent with "ICPR 2024 Resource-Limited Infrared Small Target Detection Challenge Track 2: Lightweight Infrared Small Target Detection", we use pixel-level metrics (intersection over union (IoU)) and target-level metrics

(detection rate $P_d$ and false alarm rate $F_a$) to evaluate the detection accuracy of the network [21], and use the parameters (P) and FLOPs (F) to evaluate the resource consumption of the network. Among them, the IoU and $P_d$ are used as the main detection accuracy evaluation metrics to evaluate the regional segmentation accuracy and target detection accuracy of the method, respectively. Notably, we perform statistical calculations on FLOPs according to the size of the actual input to the test network. In addition, consistent with the competition, the final performance evaluation formula of the network is as follows:

$$S_{pe} = \gamma \times S_p + (1-\gamma) \times S_e \qquad (4)$$

$$S_p = \alpha \times IoU + (1-\alpha) \times P_d \qquad (5)$$

$$S_e = 1 - \frac{P_{sub}/P_{base} + F_{sub}/F_{base}}{2} \qquad (6)$$

where $\alpha = 0.5$, $\gamma = 0.5$, $P_{base} = 2.225M$, and $F_{base} = 12.56G$. $P_{base}$ and $F_{base}$ are the benchmark parameters and FLOPs. $P_{sub}$ and $F_{sub}$ are the parameters and FLOPs of the tested method. $S_{pe}$ denotes the final *score*. In addition, $F_a$ has a constraint effect. When $F_a$ is less than $1e^{-4}$, the final result is considered valid.

**Competition details.** In "ICPR 2024 Resource-Limited Infrared Small Target Detection Challenge Track 2: Lightweight Infrared Small Target Detection", in addition to using the proposed LR-Net, we also apply test-time augmentation and adjustable sensitivity strategies [24] to the output results for post-processing. This manuscript does not involve research on related post-processing but focuses on improving network performance.

### 4.3 Ablation experiment

**Break-down Ablation.** To verify the effectiveness of each component of LR-Net, we conduct ablation experiments on the LFEA module, LFD strategy, RFT module, and SBAM. In Table 1, "Ours-w/o LFEA" removes the LFEA module and directly uses a layer of depthwise separable convolution with a kernel size of 3 and a stride of 2. "Ours-w/o LFD" directly removes the LFD strategy. "Ours-w/o RFT" removes the RFT module and directly performs feature transfer. "Ours-w/o SBAM" removes SBAM and uses the feature fusion method in the FPN.

From Table 1, first, the proposed LFEA module can greatly improve the refined segmentation effect of the network. Compared with "w/o LFEA", LR-Net achieves an increase of 8.27 in IoU (from 34.27 to 42.54). Meanwhile, the $P_d$ of "w/o LFEA" is better than that of "w/o LFD", "w/o RFT" and "w/o SBAM" due to the fact that more inaccurate detection areas will appear. The results of Fa can also illustrate this. This result further illustrates that the proposed LFEA module is simple and efficient. Secondly, compared with "w/o LFD" and "w/o SBAM", LR-Net's improvement on $P_d$ is more significant than that on IoU. This shows that for the task of infrared small target detection, fully considering and solving the problem of target loss in high-level feature maps will significantly help improve the detection rate. Thirdly, compared with "w/o RFT", LR-Net significantly improves the IoU and $P_d$, which shows that the

Table 1. Performance verification of each component of the LR-Net on the W-IRSTD dataset.

| Schemes | Components | | | | Performance Evaluation | | | | | |
|---|---|---|---|---|---|---|---|---|---|---|
| | LFEA | LFD | RFT | SBAM | IoU | $P_d$ | $F_a \times 10^{-6}$ | $P_{sub}$ | $F_{sub}$ | *Score* |
| Ours-w/o LFEA | ✗ | ✓ | ✓ | ✓ | 34.27 | 60.19 | 36.01 | 0.016 | **0.043** | 73.35 |
| Ours-w/o LFD | ✓ | ✗ | ✓ | ✓ | 42.25 | 59.43 | 21.90 | 0.018 | 0.059 | 75.10 |
| Ours-w/o RFT | ✓ | ✓ | ✗ | ✓ | 41.16 | 58.30 | 17.11 | **0.015** | 0.061 | 74.58 |
| Ours-w/o SBAM | ✓ | ✓ | ✓ | ✗ | 41.98 | 55.13 | **16.53** | 0.020 | 0.131 | 73.79 |
| **LR-Net** | ✓ | ✓ | ✓ | ✓ | **42.54** | **63.82** | 25.15 | 0.020 | 0.063 | **76.24** |

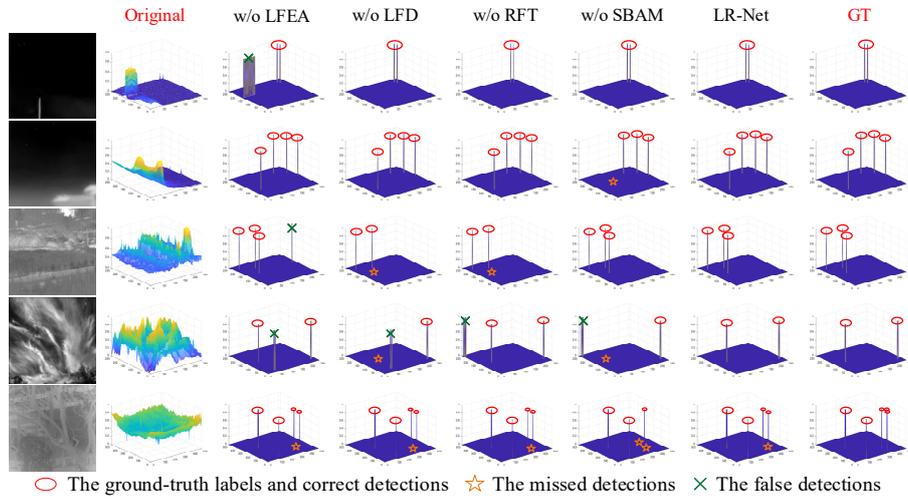

○ The ground-truth labels and correct detections   ☆ The missed detections   ✕ The false detections

**Fig. 5.** Partial test results of each component for the ablation scheme.

RFT module helps promote the refined extraction and transfer of features. Finally, from the results of the parameters and FLOPs, when the input image size is 256×256 pixels, the parameters and FLOPs of LR-Net are only 0.02 M and 0.063 G. To further intuitively demonstrate the performance of each component, we have visualized some of the test results. From Fig. 5, we can see that the final LR-Net has the best detection effect.

**Performance verification of the ECA module.** To build a robust and lightweight network, we use a lightweight ECA module in the network to explore the relationships between feature map channels. From Table 2, compared with not using the ECA module, the LR-Net with the ECA module increases by 8.07 (from 34.47 to 42.54) in IoU and 8.27 (from 55.55 to 63.82) in $P_d$. At the same time, it is worth mentioning that the parameters hardly changed after adding ECA, and FLOPs only increased by 0.002G. Secondly, by comparing the schemes after adding various attention mechanisms, the resource consumption of the ECA module is the lowest. At the same time, when the SE module [25] is used and the channel compression rate is large, its detection effect is not as good as that of the ECA module. In addition, adding the CBAM module [26] has the highest IoU, which shows that for the infrared small target detection task, the addition of the CBAM module enables the network to have a finer segmentation accuracy. However, the addition of the CBAM module introduces relatively more resource consumption, and the target detection rate is not as good as that of ECA module. Finally, a comparison of the results of different channel compression rates reveals that a channel compression rate that is too large significantly affects the performance of the network. However, the ECA module uses one-dimensional convolution operations to learn the relationships between channels and has better detection performance. Notably, since the number of channels of our initial network layer output feature map is 4, the maximum channel compression rate in the experiment is 4.

**Table 2.** Performance verification of the ECA module. *r* denotes the channel compression rate, and *k* denotes the convolution kernel size.

| Schemes | IoU | $P_d$ | $F_a \times 10^{-6}$ | $P_{sub}$ | $F_{sub}$ | *Score* |
|---|---|---|---|---|---|---|
| Ours-w/o ECA | 34.47 | 55.55 | 34.93 | **0.020** | **0.061** | 72.16 |
| Ours-w/ SE (r=4) | 41.49 | 61.43 | 18.90 | 0.029 | 0.063 | 75.28 |
| Ours-w/ SE (r=2) | 42.99 | 63.58 | 29.16 | 0.038 | 0.063 | 76.09 |
| Ours-w/ CBAM (k=3, r=4) | 41.54 | 56.65 | 25.35 | 0.029 | 0.071 | 74.08 |
| Ours-w/ CBAM (k=3, r=2) | 43.23 | 61.35 | 25.66 | 0.038 | 0.071 | 75.58 |
| Ours-w/ CBAM (k=7, r=4) | **43.99** | 61.82 | 20.56 | 0.031 | 0.104 | 75.90 |
| Ours-w/ CBAM (k=7, r=2) | 43.82 | 60.69 | **15.77** | 0.040 | 0.104 | 75.47 |
| **LR-Net** (w/ ECA) | 42.54 | **63.82** | 25.15 | **0.020** | 0.063 | **76.24** |

**Table 3.** LR-Net performance under different configurations. "Resize" denotes resizing the image to a fixed size. "Crop" denotes the window cropping method. "Chanel" denotes the channel dimension of the output feature map at each stage.

| Schemes | Configurations | | | Performance Evaluation | | | | | |
|---|---|---|---|---|---|---|---|---|---|
| | Resize | Crop | Channel | IoU | $P_d$ | $F_a \times 10^{-6}$ | $P_{sub}$ | $F_{sub}$ | *Score* |
| Resize (256) | 256 | ✗ | ×1 | 31.70 | 36.16 | 12.31 | 0.020 | 0.063 | 66.61 |
| Resize (512) | 512 | ✗ | ×1 | 40.40 | 46.63 | **11.85** | 0.020 | 0.253 | 71.03 |
| Resize (768) | 768 | ✗ | ×1 | 44.57 | 56.52 | 13.32 | 0.020 | 0.569 | 73.90 |
| Crop (256) | ✗ | 256 | ×1 | 42.54 | 63.82 | 25.15 | 0.020 | 0.063 | **76.24** |
| Crop (512) | ✗ | 512 | ×1 | 43.58 | 60.27 | 18.11 | 0.020 | 0.253 | 75.23 |
| Channel (×0.5) | ✗ | 256 | ×0.5 | 37.77 | 55.81 | 25.68 | **0.006** | **0.024** | 73.28 |
| Channel (×2) | ✗ | 256 | ×2 | 46.82 | 62.58 | 16.36 | 0.073 | 0.192 | 76.15 |
| Channel (×4) | ✗ | 256 | ×4 | **51.81** | **67.70** | 20.50 | 0.276 | 0.652 | 75.48 |

**Table 4.** Performance comparison of LR-Net and various excellent infrared small target detection methods on the W-IRSTD dataset.

| Methods | IoU | $P_d$ | $F_a \times 10^{-6}$ | $P_{sub}$ | $F_{sub}$ | *Score* |
|---|---|---|---|---|---|---|
| ACM [17] | 40.70 | 56.89 | **11.72** | 0.398 | 0.405 | 68.96 |
| ALCNet [18] | 44.96 | 58.41 | 19.97 | 0.427 | 0.381 | 70.29 |
| MLCL-Net [19] | 54.41 | 67.30 | 25.32 | 0.560 | 6.214 | 61.77 |
| ALCL-Net [15] | 54.71 | 68.25 | 18.05 | 5.668 | 6.805 | 3.51 |
| DNANet [21] | 55.62 | 69.59 | 20.26 | 4.697 | 14.282 | 0.10 |
| GGL-Net [12] | 53.94 | 72.13 | 23.45 | 8.988 | 9.856 | -39.09 |
| UIU-Net [22] | 56.24 | 69.16 | 15.32 | 50.541 | 54.501 | -595.01 |
| MSDA-Net [13] | **58.18** | **74.99** | 19.32 | 4.791 | 21.225 | -12.79 |
| **LR-Net** | 42.54 | 63.82 | 25.15 | **0.020** | **0.063** | **76.24** |

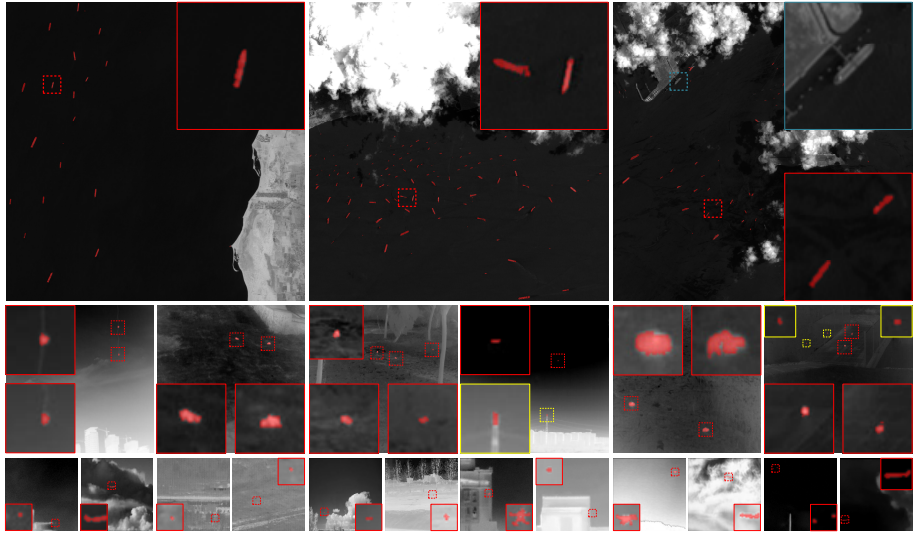

**Fig. 6.** Some results of LR-Net on the W-IRSTD dataset. The red box denotes correct detection, the blue box denotes missed detection, and the yellow box denotes false detection.

**Exploration of network details.** To further verify the rationality of the constructed LR-Net, we explore the performance of LR-Net under different configurations. From Table 3, first, compared with the resizing method, the cropping method has better detection accuracy, especially for the detection rate. This is because the samples generated by the resizing method are limited, which will cause the network to focus more on the target area with high confidence for the test samples, and thus the targets of some difficult samples will be missed. The false positive rate will also be low in this case. The comparison of $F_a$ can also illustrate this. Secondly, for the resizing strategy, "Resize(768)" is better than "Resize(256)" and "Resize(512)". The reason is that the sample size range of the used dataset is mainly in the range of 256-1024, among which the 1024×1024 size accounts for a large proportion. Thirdly, regarding the cropping strategy, "Crop(256)" is better than "Crop(512)" because "Crop(256)" can learn more samples during training. Finally, from the experimental results of the number of channels, increasing the output feature channels by a reasonable multiple significantly increases the detection accuracy. But it also results in a sharp increase in parameters and FLOPs. For the final solution, we achieve the best *score* when the cropping window is 256 and the output feature map channels of each stage are the original (4, 8, 16, 32, 64).

### 4.4    Comparison with other SOTA methods

To better verify the performance of the proposed LR-Net, we compare the performance of LR-Net with a variety of excellent infrared small target detection networks. From Table 4, first, although LR-Net is inferior to the latest methods, such as MSDA-Net, in terms of detection accuracy, LR-Net has the highest *score*. This shows that LR-Net is an efficient network that comprehensively balances detection accuracy and resource consumption. At the same time, the *score* of LR-Net are much higher than those of other excellent infrared small target detection methods. Secondly, it can be found that the parameters and FLOPs of LR-Net are only 0.020 M and 0.063 G, respectively. Compared with other methods, its lightweight effect is significant. Finally, for some of the latest excellent methods, although their detection accuracy has been improved, their resource consumption is also very high, which leads to their final low *score* or even negative *score*.

To more intuitively demonstrate the segmentation effect of LR-Net, we draw the final generated mask onto the original image for presentation. Fig. 6 shows that the segmentation effect of LR-Net is relatively accurate. Of course, there are still false detections and missed detections for some difficult samples with complex shapes and backgrounds.

## 5    Conclusion

This manuscript proposes an innovative lightweight and robust network (LR-Net) for infrared small target detection, which abandons the complex structure and achieves an effective balance between detection accuracy and resource consumption. Specifically, we construct an LFEA module, which can effectively extract features of infrared small target images with little resource consumption. By superimposing LFEA modules, a multi-stage basic network for infrared small target feature extraction can be constructed. At the same time, to solve the problem of target loss in high-level feature maps, on the one hand, we propose an LFD strategy, which can effectively supplement target information. On the other hand, we introduce SBAM to utilize low-level features to guide the constraints and fusion of high-level features, thereby improving the detection capabilities. In addition, we build a simple RFT module, which can improve the network's feature refinement extraction capability with little resource consumption. Moreover, for the specific training and testing strategies, we abandon the traditional method of directly resizing the image to a fixed size and adopt a new cropping method. This method shows better robustness for datasets with samples of multiple sizes. Extensive experimental results show that our method achieves the best comprehensive performance.